\documentclass{article}
\usepackage[utf8]{inputenc}
\usepackage{graphicx}
\usepackage{multirow}
\usepackage{booktabs}
\usepackage{hyperref}
\usepackage{caption}
\usepackage{subcaption}
\usepackage{authblk}
\usepackage{amsmath}

\usepackage[a4paper, total={6in, 9in}]{geometry}

\usepackage{xcolor}

\title{PTT5: Pretraining and validating the T5 model on Brazilian Portuguese data}
\author[1]{Diedre Carmo}
\author[1]{Marcos Piau}
\author[1,2]{Israel Campiotti}
\author[1,2,3]{Rodrigo Nogueira}
\author[1,2]{Roberto Lotufo}

\affil[1]{School of Electrical and Computing Engineering, UNICAMP}
\affil[2]{NeuralMind Inteligência Artificial}
\affil[3]{David R. Cheriton School of Computer Science, University of Waterloo}
\date{\textbf{Source code: \url{https://github.com/unicamp-dl/PTT5}\\August 2020}}

\begin{document}

\maketitle

\begin{abstract}
    In natural language processing (NLP), there is a need for more resources in Portuguese, since much of the data used in the state-of-the-art research is in other languages. In this paper, we pretrain a T5 model on the BrWac corpus, an extensive collection of web pages in Portuguese, and evaluate its performance against other Portuguese pretrained models and multilingual models on three different tasks. We show that our Portuguese pretrained models have significantly better performance over the original T5 models. Moreover, we demonstrate the positive impact of using a Portuguese vocabulary.
    Our code and models are available at \url{https://github.com/unicamp-dl/PTT5}.
\end{abstract}

\section{Introduction}

Pretrained language models have been employed successfully in various NLP tasks~\cite{devlin2018bert, raffel2019exploring, liu2019roberta}. Recent works demonstrate that monolingual pretrained models perform better on tasks in that same language than models pretrained on multilingual corpora~\cite{souza2019Portuguese, virtanen2019multilingual,martin2019camembert,delobelle2020robbert,vries2019bertje,Canete2020beto,polignano2019alberto,baly2020arabert,kuratov2019rubert,nguyen2020phobert}.  

For Portuguese tasks, BERT models have already shown improved performance when pretrained on a Portuguese corpus~\cite{souza2019Portuguese}. One of the motivations to perform a similar pretraining but using the T5 model~\cite{raffel2019exploring} is its capability to generate text. Thus, it can perform tasks that a BERT model cannot, such as summarization, abstractive question answering, and translation.

In this work, we improve the original T5 model on Portuguese language tasks by pretraining it on BrWac~\cite{wagner2018brwac}, a large corpus of web pages in Brazilian Portuguese. We call this model PTT5. We validate our pretraining on Portuguese tasks of sentence entailment prediction and named entity recognition, and show that monolingual pretraining significantly improves the model's performance.

\section{Data}

We use three Brazilian Portuguese datasets: BrWac~\cite{wagner2018brwac} for pretraining, and ASSIN 2~\cite{real2020assin} and HAREM for fine-tuning and evaluating our pretrained models. 

\subsection{BrWac}
The BrWac corpus~\cite{wagner2018brwac} was built after crawling more than 60 million web pages, filtered down to 3.5 million pages after applying quality control filters, resulting in a dataset with 2.7 billion tokens.

We construct input examples for the pretraining task by concatenating sentences until we reach 512 words. If the last sentence does not fit in the 512 words, we move it to the next example, i.e., we do not split sentences. Any sentence larger than 512 words, which is rare, is truncated to 512 words and added as a separate input example, truncated to 512 words. We use the ftfy library~\cite{speer-2019-ftfy} to fix encoding problems. The resulting dataset has 15.6 GB of text and a mean of 360 words per document, as shown in Table~\ref{tab:brwac}. We use Python's split function to count words.

\begin{table}[ht]
\centering
\begin{tabular}{lr}
\toprule
\multicolumn{2}{c}{\textbf{Processed BrWac Statistics}} \\\midrule
Number of documents               & 7,361,359           \\
Total number of words             & 2,656,275,093  \\
Mean number of words per doc               & 360 $\pm$ 169 (std) \\\bottomrule
\end{tabular}
\caption{Statistics of BrWac text after pre-processing.}
\label{tab:brwac}
\end{table}

These input examples go through a tokenizer that can vary from using the original T5 vocabulary or our custom Portuguese vocabulary (Section~\ref{sec:vocab}).

\subsection{ASSIN 2}
ASSIN 2~\cite{real2020assin} consists of two Portuguese tasks: semantic similarity and entailment prediction. In the semantic similarity task, given a pair of sentences, a model has to predict a number between 1 and 5, representing how semantically close the two sentences are. The entailment prediction task consists of classifying if one sentence implies the other. It is thus a binary classification task whose classes are ``entail'' or ``none''. The dataset consists of short sentence pairs, with 6500 pairs for training, 500 for validation, and 2448 for testing.

\subsection{HAREM}

HAREM \cite{santos-etal-2006-harem} is a collection of two Portuguese datasets for Named Entity Recognition (NER). The first dataset, called First HAREM, consists of 129 documents with a total of 4151 entities on the \textit{selective} scenario and 5017 for the \textit{total} scenario. Those two scenarios differs only on the quantity of annotated classes. The \textit{total} scenario contains 10 classes (Location, Person, Organization, Value, Date, Title, Thing, Event, Abstraction and Other), while the \textit{selective} scenario contains 5 classes (Person,
Organization, Location, Value and Date). The second dataset, called MiniHAREM, is composed of 128 documents, 3642 and 3018 entities for the \textit{total} and \textit{selective} scenarios, respectively. In this work, we focus only on the \textit{selective} scenario. We separate 7 percent of the First HAREMs documents for validation and the remaining for training. All MiniHAREM is used as a test set. We refer the reader to the original paper for more information \cite{santos-etal-2006-harem}.

\section{Methodology}
We now describe our methodology, including creating the custom Portuguese vocabulary, unsupervised pretraining, and fine-tuning and evaluation on ASSIN 2 and HAREM.

\subsection{Portuguese Vocabulary}
\label{sec:vocab}

The original T5 vocabulary uses the SentencePiece library~\cite{kudo2018sentencepiece} using English, German, French, and Romanian web pages from Common Crawl.\footnote{\url{http://commoncrawl.org/}}

We use a similar procedure to create our Portuguese vocabulary: we train a SentencePiece model on a corpus of 2 million sentences randomly chosen from the Portuguese Wikipedia. We use the Unigram language model as in Kudo~\cite{kudo2018subword} and a predetermined vocabulary size of 32,000 wordpieces.

We use the same control tokens (padding, end-of-sequence, and unknown) and vocabulary size of the original T5 to start pretraining from the original T5 checkpoints without significant changes in the model architecture and overall pretraining process.

\subsection{Unsupervised Pretraining}
\label{sec:pre_train}
The unsupervised pretraining was performed with a denoising objective, which can be implemented in a few different ways. The main idea is to train the model in an unsupervised way, feeding the model with corrupted versions of the original token sequence, and training it to reconstruct the original sequence~\cite{raffel2019exploring}.

In all pretraining experiments, we use one of the strategies explored in the original T5 paper: each token in the input sequence has a predefined probability of being replaced by a mask token. The model is fed with this masked token sequence, and trained to produce the original sequence. For example, given the input sequence \textit{``Que $<$M$>$ para $<$M$>$ sobre o $<$M$>$ PTT5!"}, the model is trained to produce the sequence \textit{``Que belo dia para aprendermos sobre o maravilhoso PTT5!"}, where ``$<$M$>$" is a mask token. Note that this is an illustrative example. In practice, the tokens in the sentence are subword units.

In the pretraining experiments, we use the cross-entropy loss as a cost function and the Adafactor~\cite{shazeer2018adafactor} optimizer. Training always starts from the corresponding original T5 checkpoints released by Raffel et al.~\cite{raffel2019exploring}. 
We use Google Cloud TPU v3-8's and T5's official implementation in TensorFlow.\footnote{\url{https://github.com/google-research/text-to-text-transfer-transformer}}

In addition to updating all model weights during pretraining, we also experiment with updating only the vocabulary embeddings and freezing the remaining weights. The model then has fewer weights to be learned during pretraining, which leads to faster convergence times. Hence, this could be a more economical pretraining strategy than the widely adopted strategy of pretraining the whole model.

\subsection{ASSIN 2 Training and Validation}

For the ASSIN 2 tasks, the input to T5 is formatted as:

\begin{equation}
\text{ASSIN sentence1: }\texttt{[S1]} \texttt{[eos]}\text{sentence2: }\texttt{[S2]} \texttt{[eos]}
\end{equation}

\noindent where \texttt{[S1]} and \texttt{[S2]} corresponds to the strings of the sentences 1 and 2, respectively, and \texttt{[eos]} is the end-of-sequence token. We experimented with two ways for producing the scores in the sentence similarity task: For the first approach, we follow Raffel et al.'s strategy for regression tasks and train the model to output the literal string representing the score. We limit this generation to 5 tokens, which is then converted to a floating-point number. In the second approach, we feed the mean over the sequence length of the last hidden state of T5's encoder to a linear layer with a sigmoid activation. Since scores are between 1 and 5, we rescale the scalar output $y$ of the sigmoid layer by doing $4y + 1$.
The loss function consists of Mean Square Error (MSE), which is also one of the the similarity tasks' main metrics.
For the entailment task, we feed the last hidden state of the T5 encoder to the linear layer with two neurons in the output followed by a softmax. We fine-tune the models using the cross-entropy loss with the RAdam~\cite{liu2019variance} optimizer.

\subsection{HAREM Training and Validation}

For the Named Entity Recognition task we feed the model with an input using the following format:
\begin{equation}
\text{Recognize Entities: }\texttt{[S]} \texttt{[eos]},
\end{equation}
\noindent where \texttt{[S]} is the sentence string.
As output we expect that each entity is followed by its class. For instance, given the input ``Recognize Entities: John lives in New York", the target output would be ``John [Person] lives in [Other] New York [Local]", where ``[Other]" is used to label all out of context tokens. The example is given in English for demonstration purpose only as HAREM is a dataset in Portuguese. This approach allows us to recognize entities in a sentence using an sequence to sequence model, without any modification in its architecture. Training is done through minimization of cross-entropy loss for each token using the AdamW optimizer~\cite{loshchilov2017decoupled}.


\section{Experiments and Discussion}
Our experiments comprise two main phases: unsupervised pretraining on the BrWac corpus, and fine-tuning and evaluating on ASSIN 2 and HAREM tasks. 

\subsection{Unsupervised Pretraining}

\begin{figure}[ht]
\centering
\begin{subfigure}{.49\textwidth}
  \centering
  \includegraphics[width=\textwidth]{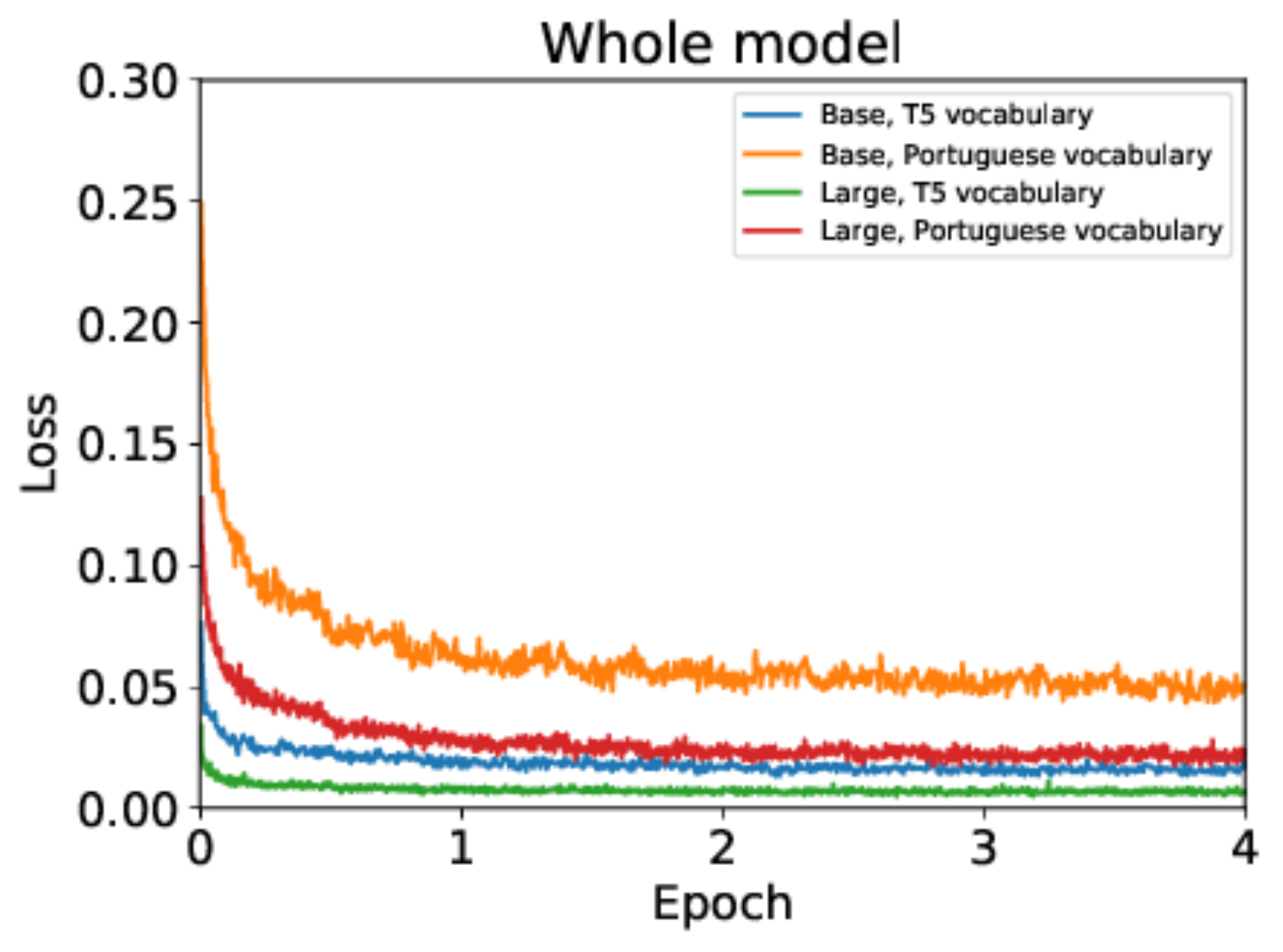}
  \caption{}
  \label{fig:pretraing_all_weights}
\end{subfigure}%
\begin{subfigure}{.49\textwidth}
  \centering
  \includegraphics[width=\textwidth]{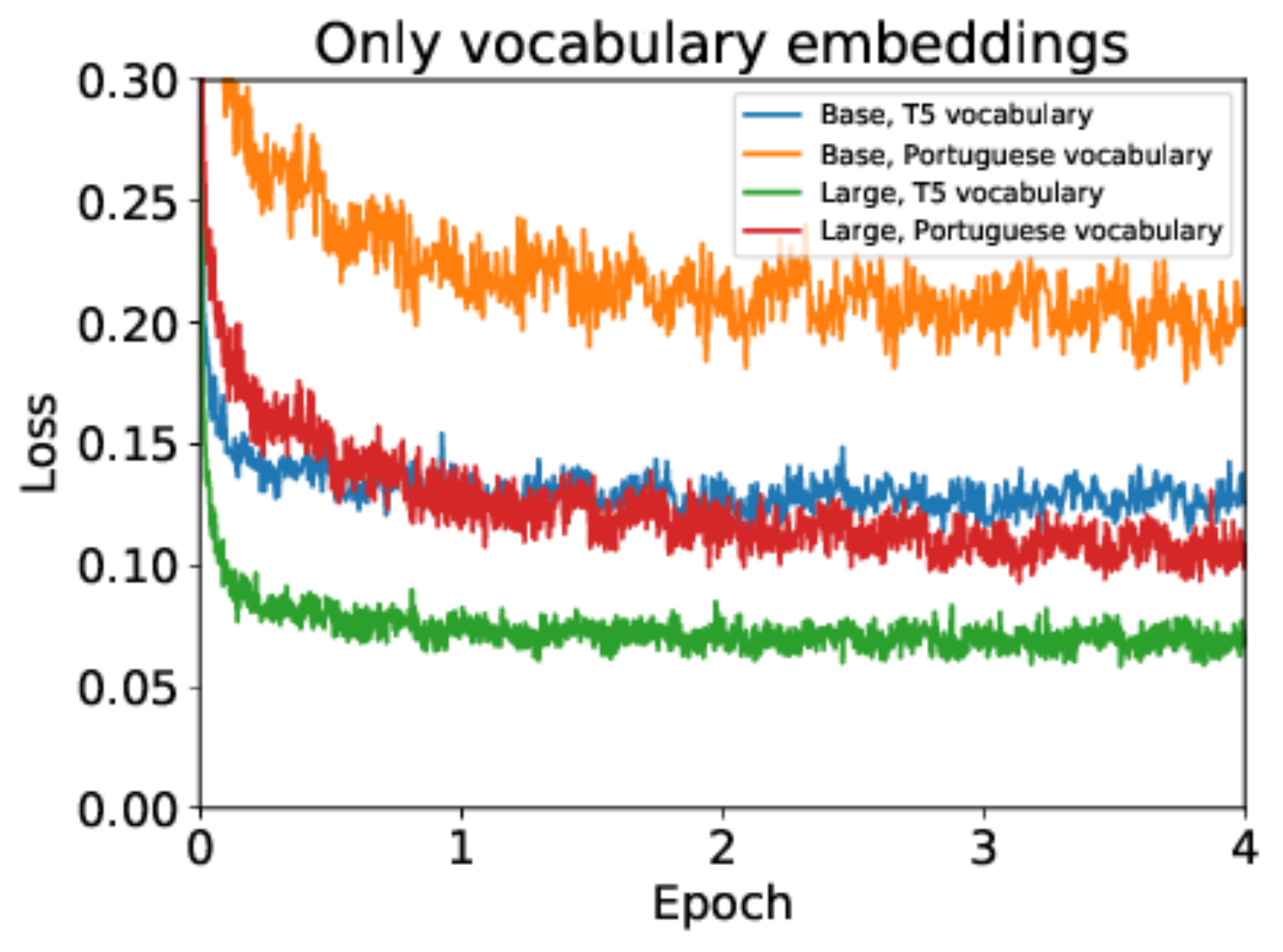}
  \caption{}
  \label{fig:pretraing_emb_only}
\end{subfigure}%

\caption{Pretraining cross-entropy loss for different model sizes and vocabularies. a) training the whole model; b) training vocabulary embeddings only.}
\label{fig:pretrain_loss_comp}
\end{figure}

The experiments with unsupervised pretraining (Figure~\ref{fig:pretrain_loss_comp}) were conducted following the methodology described in Section~\ref{sec:pre_train}. The corruption rate for the input tokens was 15\%, which is the same used by the original T5 model. Input and output sequence token lengths follow T5's maximum of 512. Longer sequences are truncated, and shorter sequences are padded with a special padding token. The learning rate was constant and equal to 0.003. All models were pretrained for four epochs.

Table~\ref{tab:pretrain_config} shows a summary of all pretraining experiments performed. Each row represents a specific combination of model size, batch size, and pretraining strategy (whole model vs. vocabulary embeddings only) and comprises two experiments: one for multilingual vocabulary and another for the Portuguese vocabulary (see Section~\ref{sec:vocab} for more details).

\begin{table}[b]
\centering
\resizebox{\columnwidth}{!}{%
\begin{tabular}{ccccc}
\hline
\multirow{2}{*}{\textbf{Size}} & \multirow{2}{*}{\textbf{Batch size}} & \textbf{Trainable parameters}      & \textbf{Hours/epoch}               \\
                               &                                      & \textbf{(all weights / emb. only)} & \textbf{(all weights / emb. only)} \\ \hline
Small & 256 & 60M / 16M  & 3.3 / 3.4   \\
Base  & 128 & 220M / 25M & 8.4 / 6.6   \\
Large & 64  & 740M / 32M & 39.5 / 32.3 \\
Large & 128 & 740M / (not performed)  & 38.9 / (not performed)  \\ 
\hline
\end{tabular}
}
\caption{\label{tab:pretrain_config} Batch size and approximated training time per epoch on a TPU v3-8 for each PTT5 model size (total amount of trainable parameters) on unsupervised pretraining experiments. We show numbers for pretraining the whole model versus pretraining the vocabulary embeddings only.}
\end{table}

As expected, larger models show lower loss levels. Using the Portuguese vocabulary resulted in higher loss values for the same model size. A hypothesis for the cause of this behavior is that, since the weights were initialized using checkpoints from models trained with the original T5 vocabulary, the model has to adapt to the new Portuguese vocabulary. Evidence for this is the initial high loss for models starting with the Portuguese vocabulary. 

For base and large models, we observed a decrease in approximately 20\% on time spent per epoch when training only the vocabulary embeddings versus training the whole model.

\begin{table}[ht]
\centering

\begin{tabular}{lrrrr}
\toprule
\multicolumn{5}{c}{\textbf{ASSIN 2 Tasks Test Results}} \\\midrule
\textbf{Team/Method}       & \textbf{Pearson}       & \textbf{MSE}      & \textbf{Accuracy} & \textbf{F1}  \\\midrule
Deep Learning Brasil~\cite{rodrigues2020assin2-dlb} & 0.785                  & 0.59              &  88.3 & 88.3\\
IPR~\cite{rodrigues2020assin2-ipr} & 0.826                  & 0.52              &  87.6 & 87.6\\
Stilingue~\cite{fonseca2020assin2-stilingue}                 & 0.817                  & \textbf{0.47}     &  86.6 & 86.6\\
mBERT ~\cite{souza2019Portuguese} & 0.809                  & 0.58              &  86.8 & 86.8\\
BERTimbau Base~\cite{souza2019Portuguese} & 0.836                  & 0.58              &  89.2 & 89.2\\
BERTimbau Large~\cite{souza2019Portuguese} & \textbf{0.852}         & 0.50              &  \textbf{90.0} & \textbf{90.0}\\
\toprule
T5 Small                   & 0.757                  & 0.61              & 83.9 & 83.9\\
T5 Base                    & 0.780                  & 0.60              & 82.3 & 82.1\\
T5 Large                   & 0.780                  &  0.57             & 83.5 & 83.2\\
\midrule
PTT5 Small, T5 vocab        & 0.776                  & 0.65              & 84.7 & 84.7\\
PTT5 Base, T5 vocab         & 0.793                  & 0.63              & 84.4 & 84.4\\
PTT5 Large, T5 vocab        & 0.798                  & 0.59              & 87.5 & 87.4\\
\midrule
PTT5 Small, PT vocab        & 0.811                  & 0.51              & 85.8 & 85.8\\
PTT5 Base, PT vocab         & \textbf{0.829}         & \textbf{0.47}     & \textbf{88.6} & \textbf{88.5}\\
PTT5 Base, PT vocab, emb. only & 0.762         & 0.67     & 85.1 & 85.0\\
PTT5 Large, PT vocab       &  0.819          &    0.53   & 88.0 & 88.0 \\\bottomrule
\end{tabular}
\caption{\label{tab:test} Test results on ASSIN 2 using the official evaluation code.}
\end{table}

\subsection{ASSIN 2 Experiments}

Here we describe our experiments in our target task: ASSIN 2.
Unless noted otherwise, the learning rate for fine-tuning is 0.0001; the batch sizes are 32, 2, and 1 for the small, base, and large models, respectively. The small batch sizes for the base and large models are due to memory limitations in our GPU. We found that 128 sequence length is enough to accommodate ASSIN 2's sentence pairs by looking at the tokenized training and validation data. The maximum number of epochs in the reported experiment plots is 50. We used a patience of 5 epochs for the similarity task and 10 epochs for the entailment task.

Table~\ref{tab:test} shows results on the test set using the official ASSIN 2 evaluation script. We compare with fine-tuning from the original T5 weights and fine-tuning from PTT5. We also compare our results to BERTimbau, a BERT model also pretrained on BrWac~\cite{souza2019Portuguese}, mBERT, a multilingual training of BERT, and the top models from the official ASSIN 2 leaderboard.\footnote{\url{https://sites.google.com/view/ASSIN 2}}.

In general, PTT5 Base achieves competitive performance with BERTimbau, with the Portuguese vocabulary largely contributing to the results. Despite the largest BERT model achieving the best performance (BERTimbau Large), our PTT5 Base was better than PTT5 Large. PTT5 Base achieves top MSE, which can be due to optimization with MSE loss. It is noticeable how ASSIN 2's test dataset is different from its validation set. 
PTT5 consistently outperforms the original T5 model on both tasks. This result aligns with our initial hypothesis that Portuguese denoise pretraining might improve performance when fine-tuning on Portuguese tasks. The use of the custom Portuguese vocabulary also consistently improved results. 

\subsubsection{Ablation: Output Strategy}

\begin{figure}[p]
\centering
\begin{subfigure}{.49\textwidth}
  \centering
  \includegraphics[width=\textwidth]{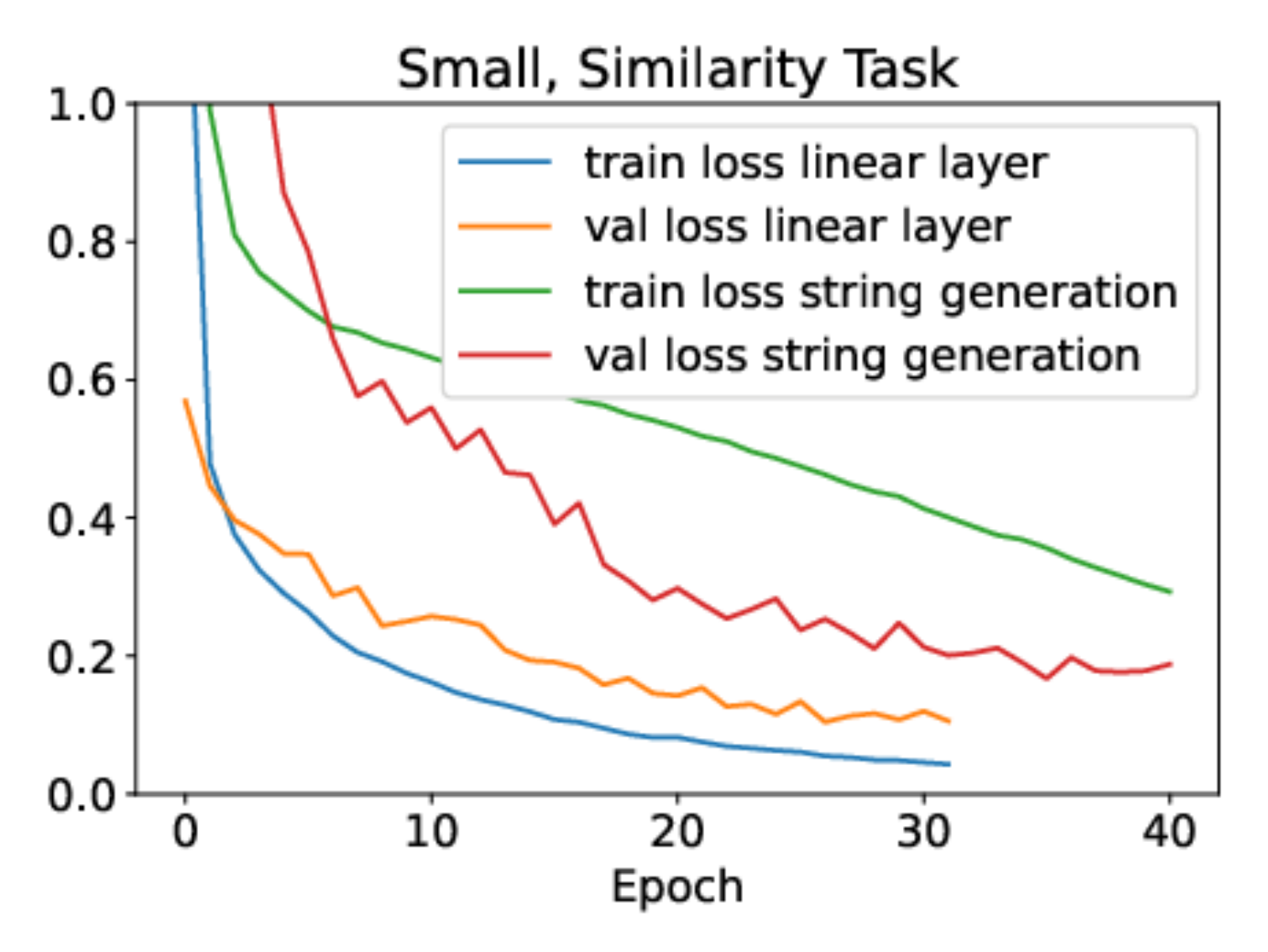}
  \caption{}
  \label{fig:small_long_gen}
\end{subfigure}%
\begin{subfigure}{.49\textwidth}
  \centering
  \includegraphics[width=\textwidth]{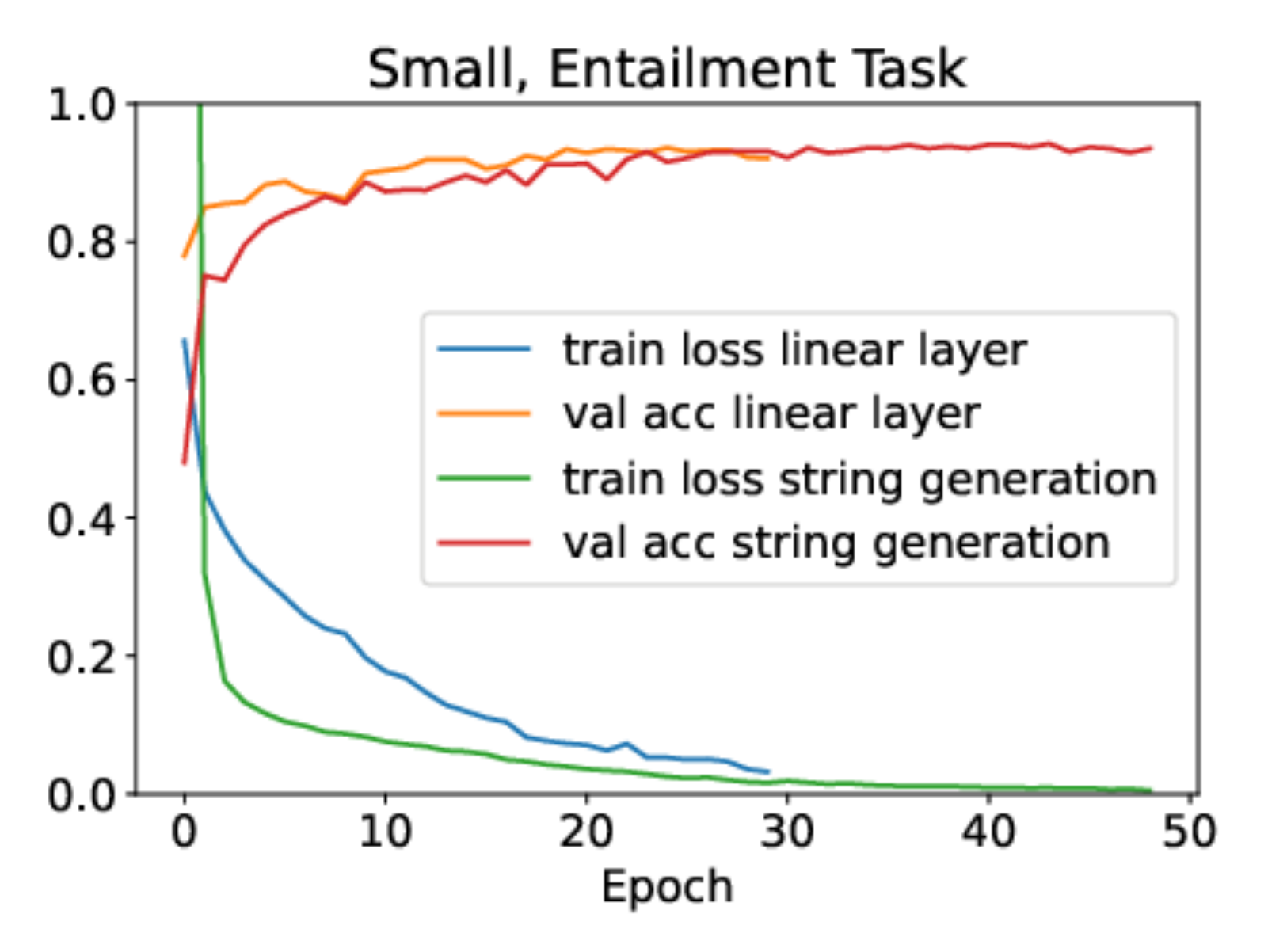}
  \caption{}
  \label{fig:small_entail_long_gen}
\end{subfigure}%

\begin{subfigure}{.49\textwidth}
  \centering
  \includegraphics[width=\textwidth]{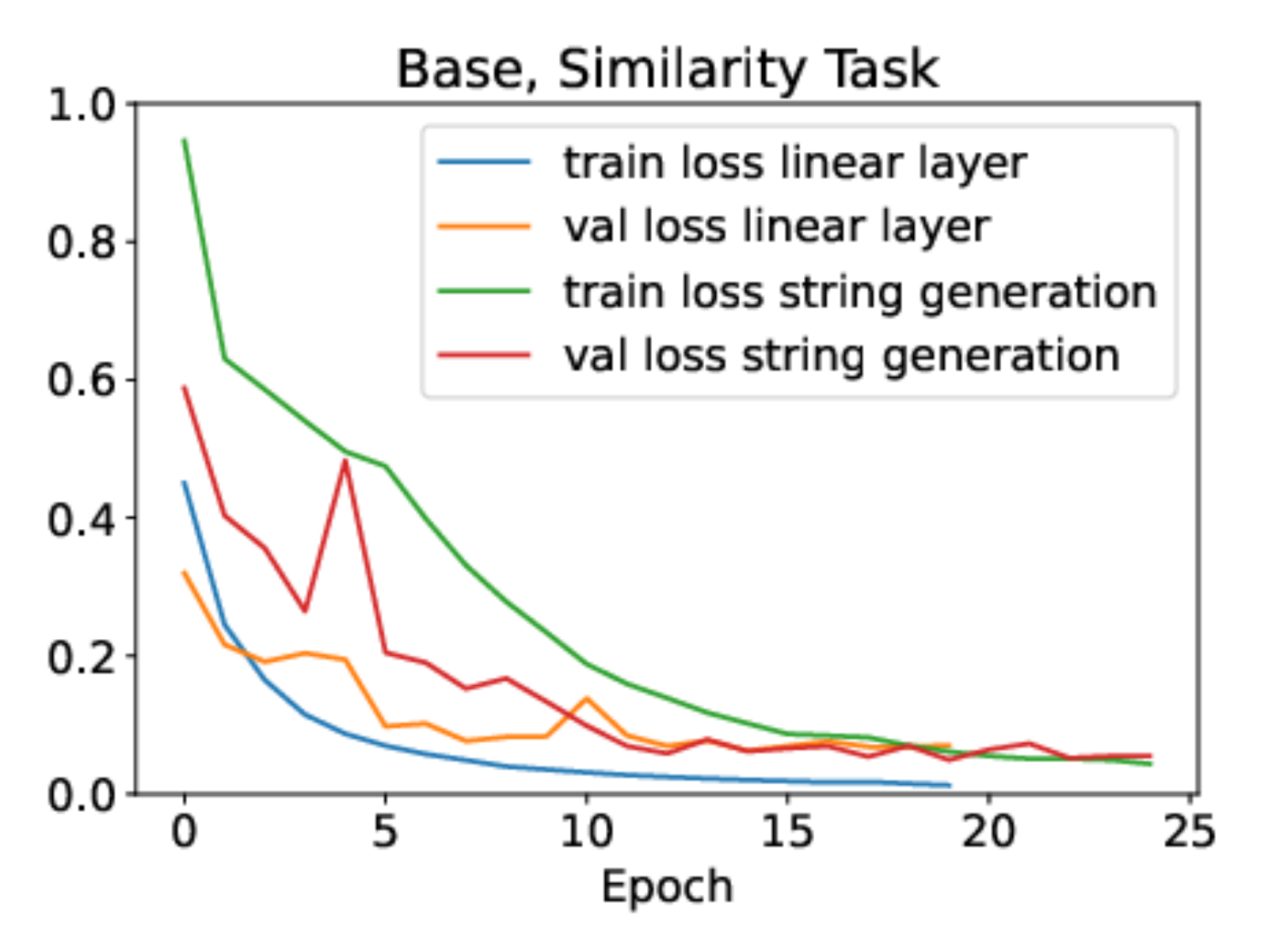}
  \caption{}
  \label{fig:base_long_gen}
\end{subfigure}
\begin{subfigure}{.49\textwidth}
  \centering
  \includegraphics[width=\textwidth]{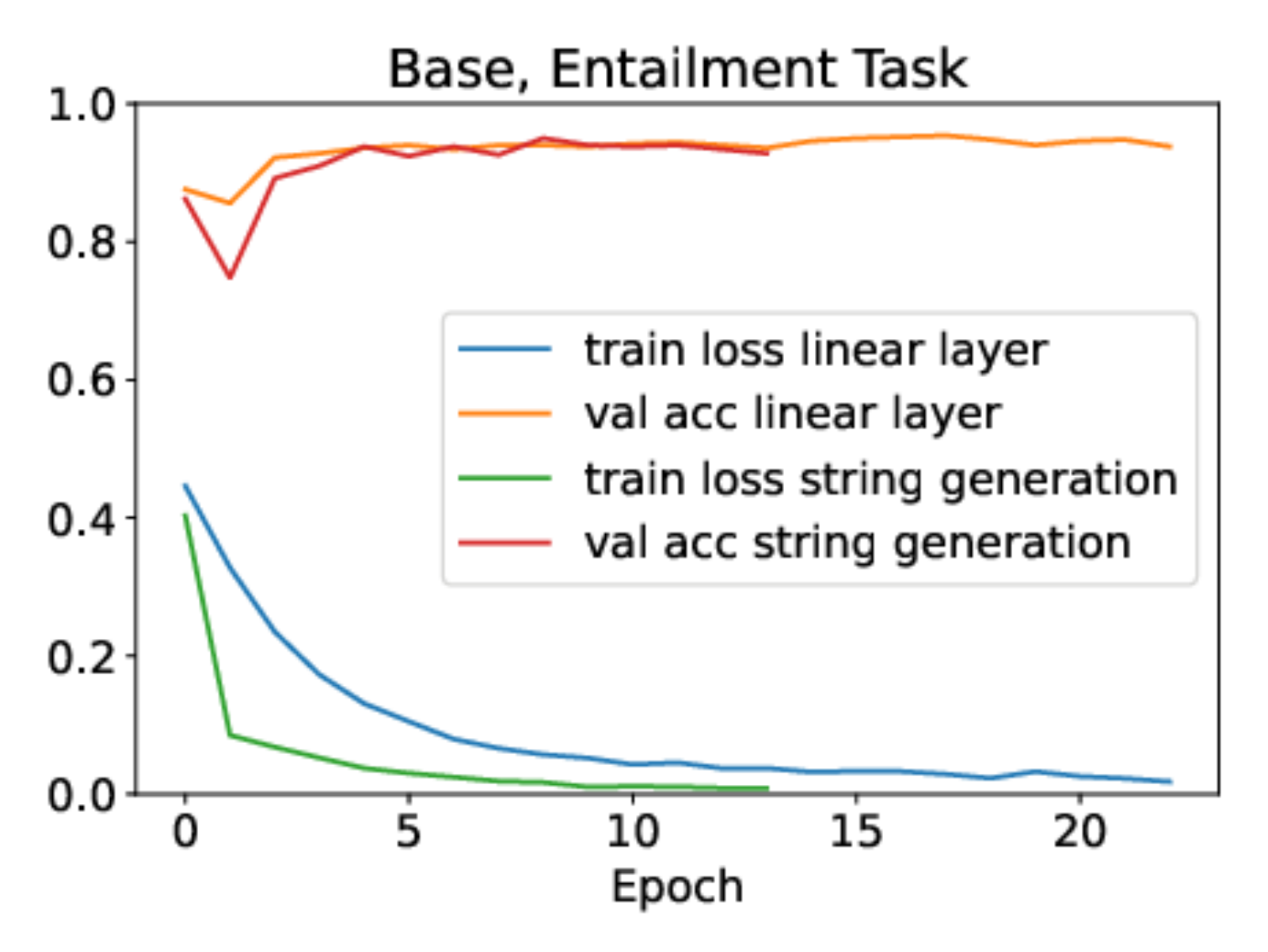}
  \caption{}
  \label{fig:base_entail_long_gen}
\end{subfigure}

\caption{Training and validation curves comparing the string generation and linear layer approaches on the similarity and entailment tasks, starting from T5 small (a and b) and T5 base weights (c and d).}
\label{fig:long_vs_gen}
\end{figure}

Figure~\ref{fig:long_vs_gen} compares the validation loss in semantic similarity task and the validation accuracy in the entailment task between the two output strategies: string generation and linear layer over the last hidden states. For the string generation approach,accuracy is used in Figures~\ref{fig:small_entail_long_gen} and  \ref{fig:base_entail_long_gen}. These experiments used the original T5 vocabulary and weights. We use the linear layer approach in all other experiments due to its faster convergence and higher stability. 

\subsubsection{Ablation study: Portuguese Vocabulary vs. Original T5 Vocabulary}

\begin{figure}[p]
\centering
\begin{subfigure}{.48\textwidth}
  \centering
  \includegraphics[width=\textwidth]{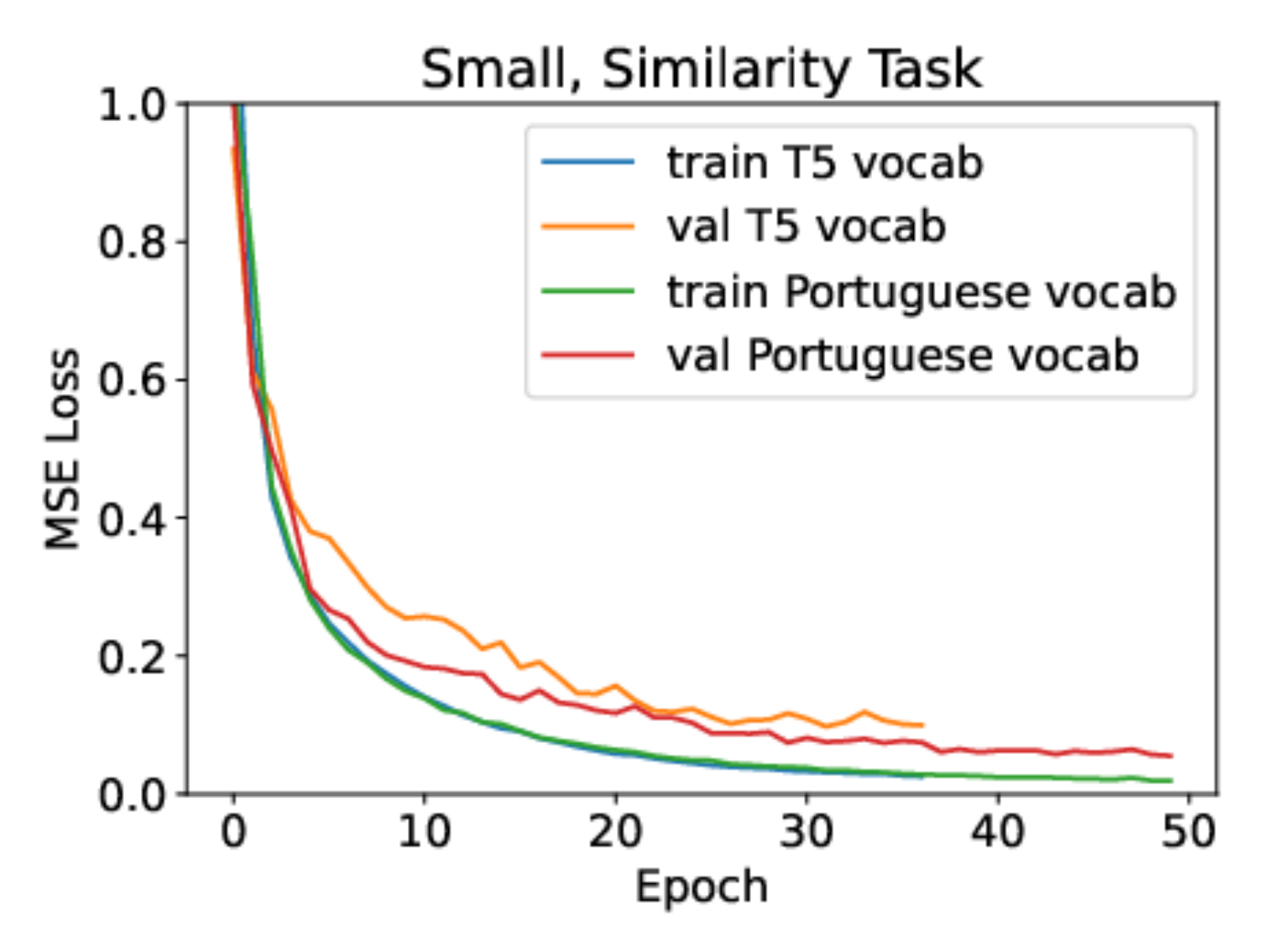}
  \caption{}
  \label{fig:small_vocab}
\end{subfigure}%
\begin{subfigure}{.48\textwidth}
  \centering
  \includegraphics[width=\textwidth]{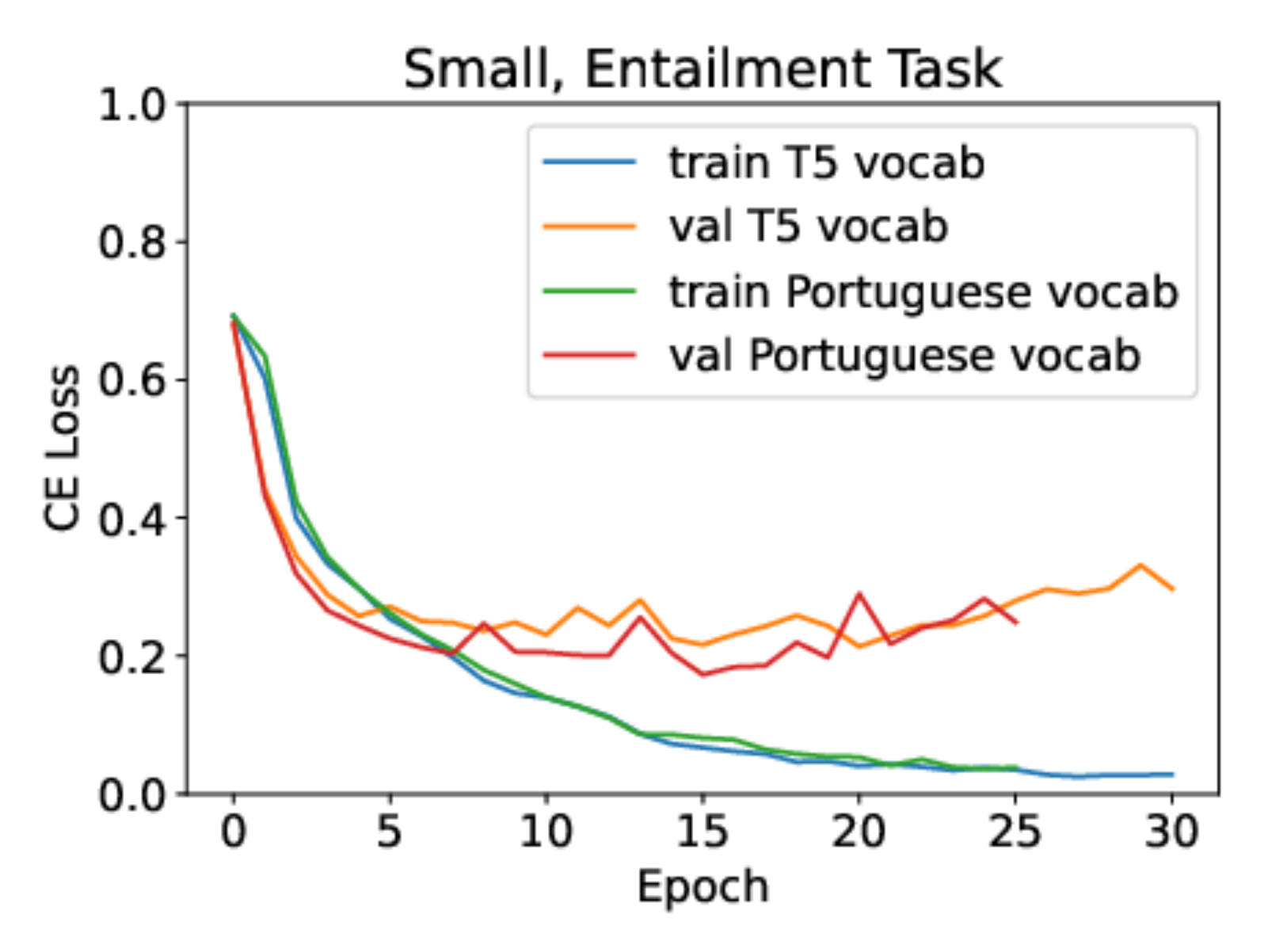}
  \caption{}
  \label{fig:vocab_vs_entail}
\end{subfigure}%

\begin{subfigure}{.48\textwidth}
  \centering
  \includegraphics[width=\textwidth]{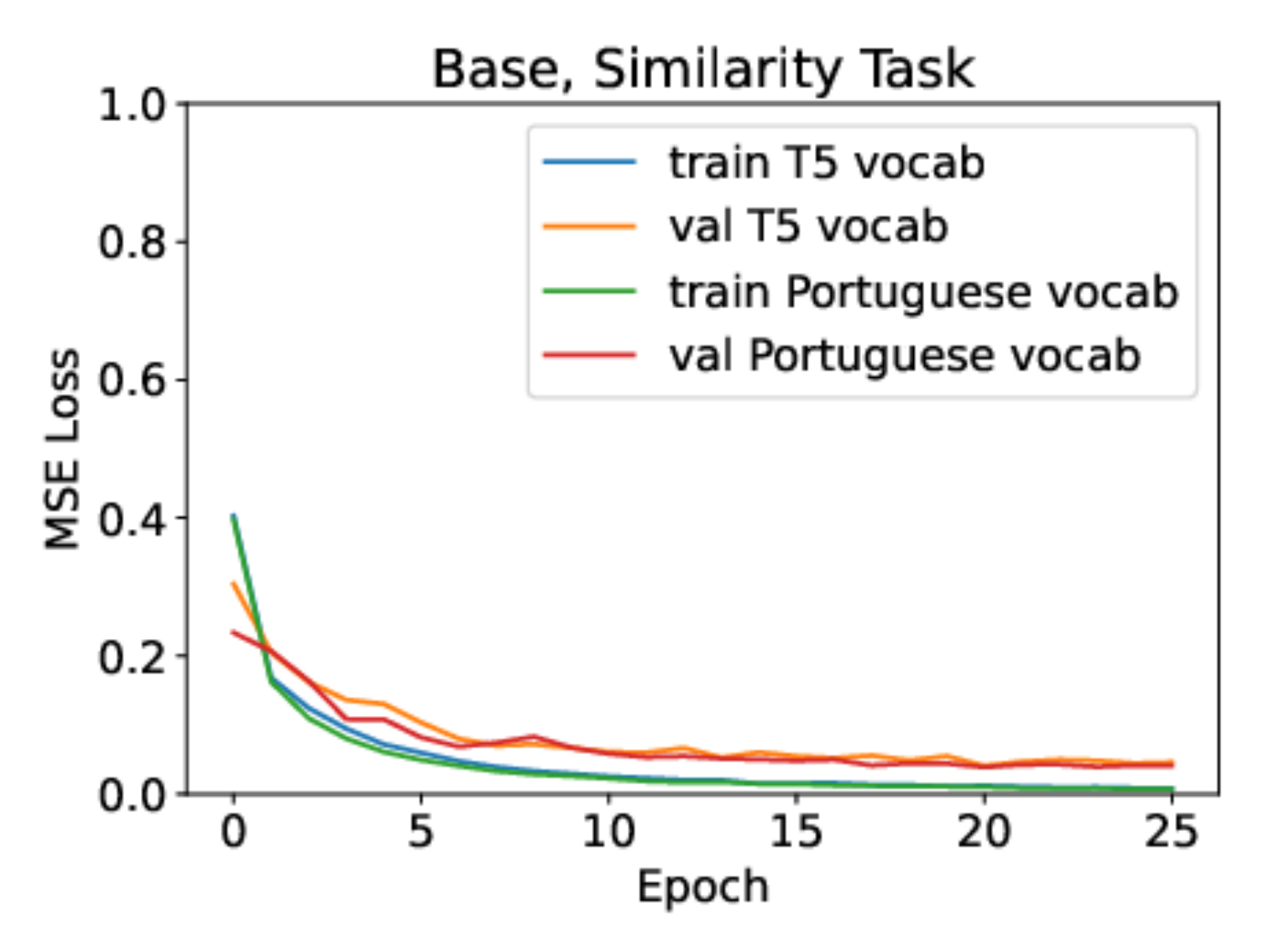}
  \caption{}
  \label{fig:base_vocab}
\end{subfigure}
\begin{subfigure}{.48\textwidth}
  \centering
  \includegraphics[width=\textwidth]{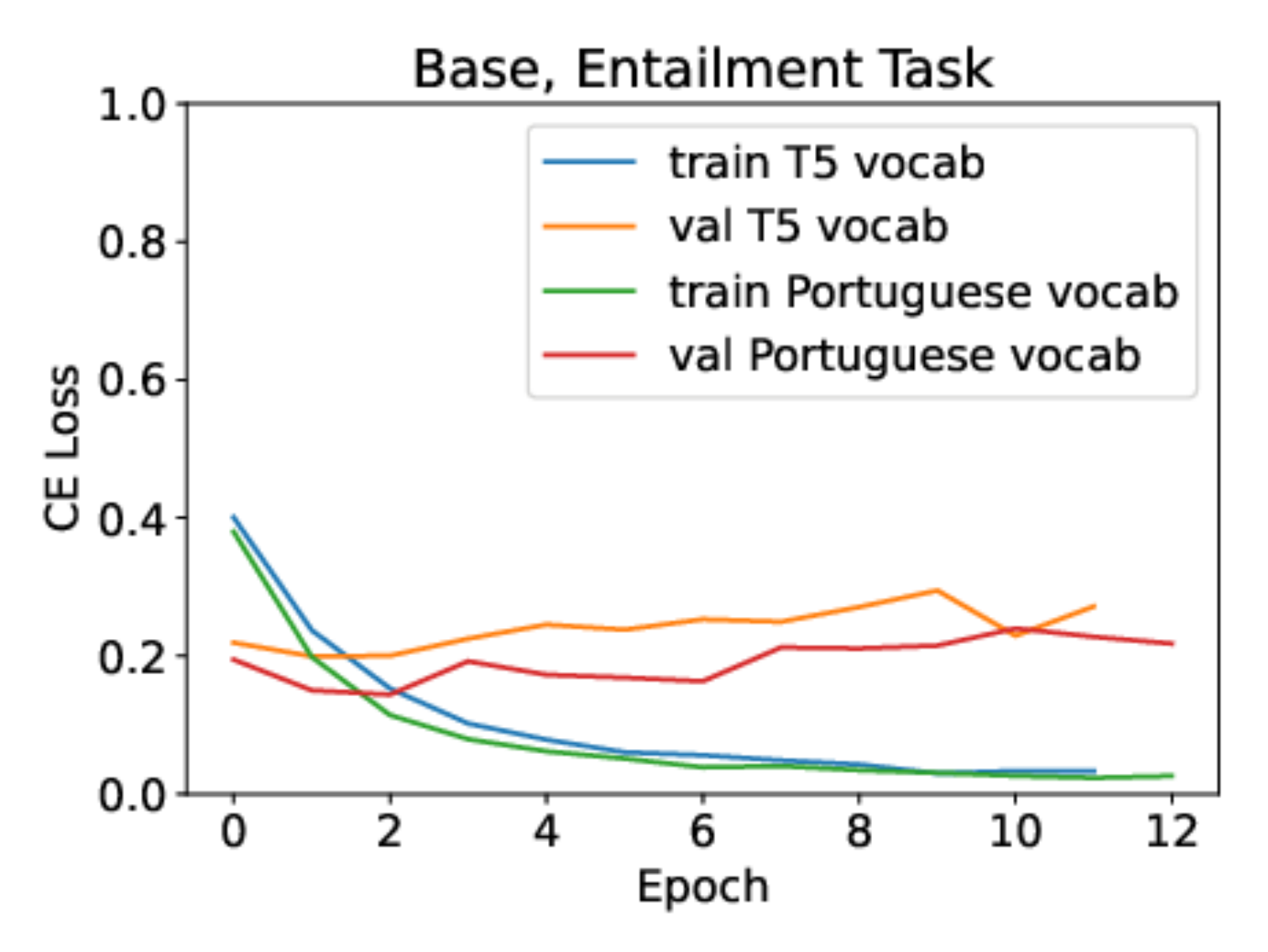}
  \caption{}
  \label{fig:base_vocab_vs_entail}
\end{subfigure}

\begin{subfigure}{.48\textwidth}
  \centering
  \includegraphics[width=\textwidth]{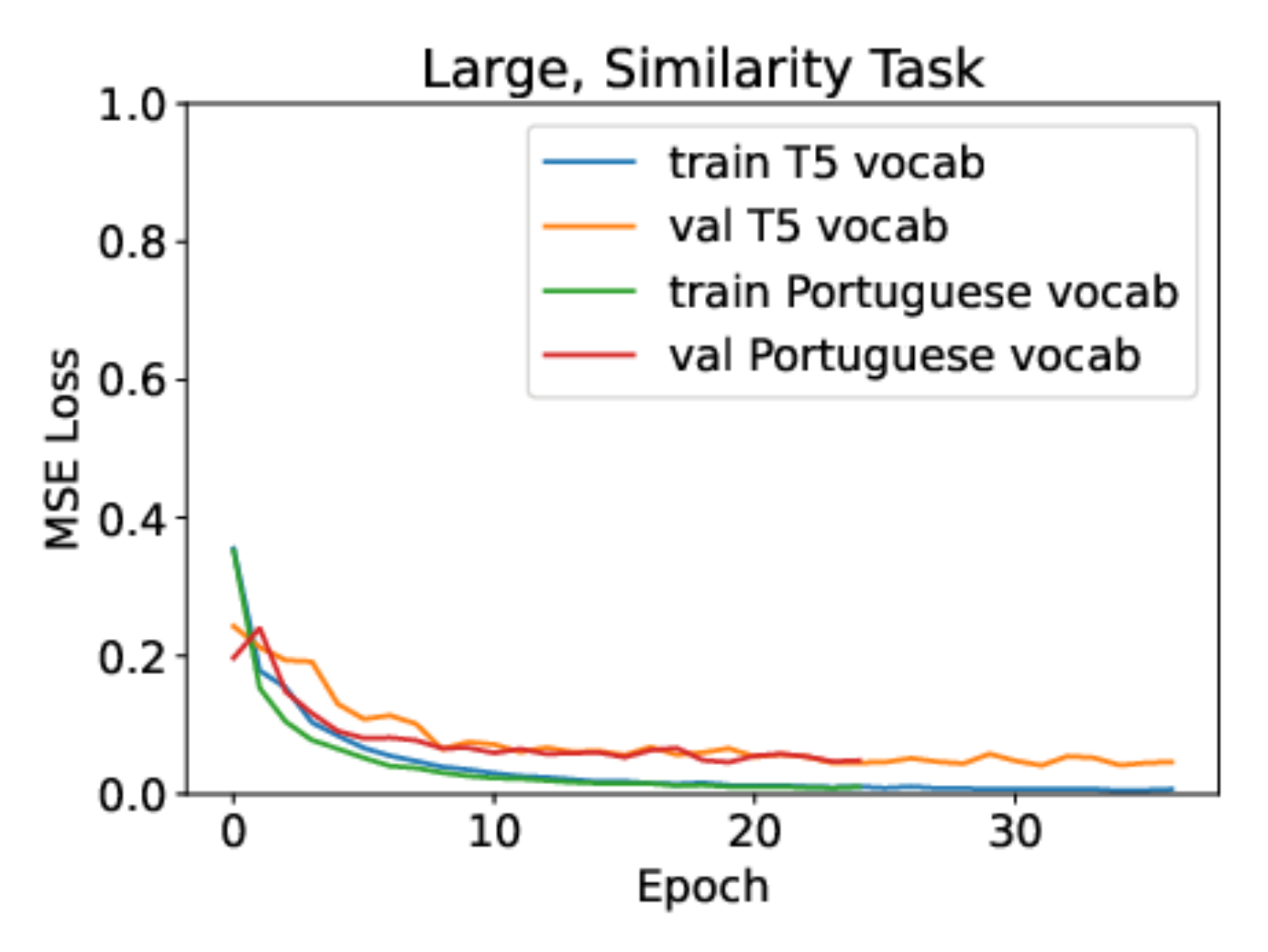}
  \caption{}
  \label{fig:large_vocab}
\end{subfigure}
\begin{subfigure}{.48\textwidth}
  \centering
  \includegraphics[width=\textwidth]{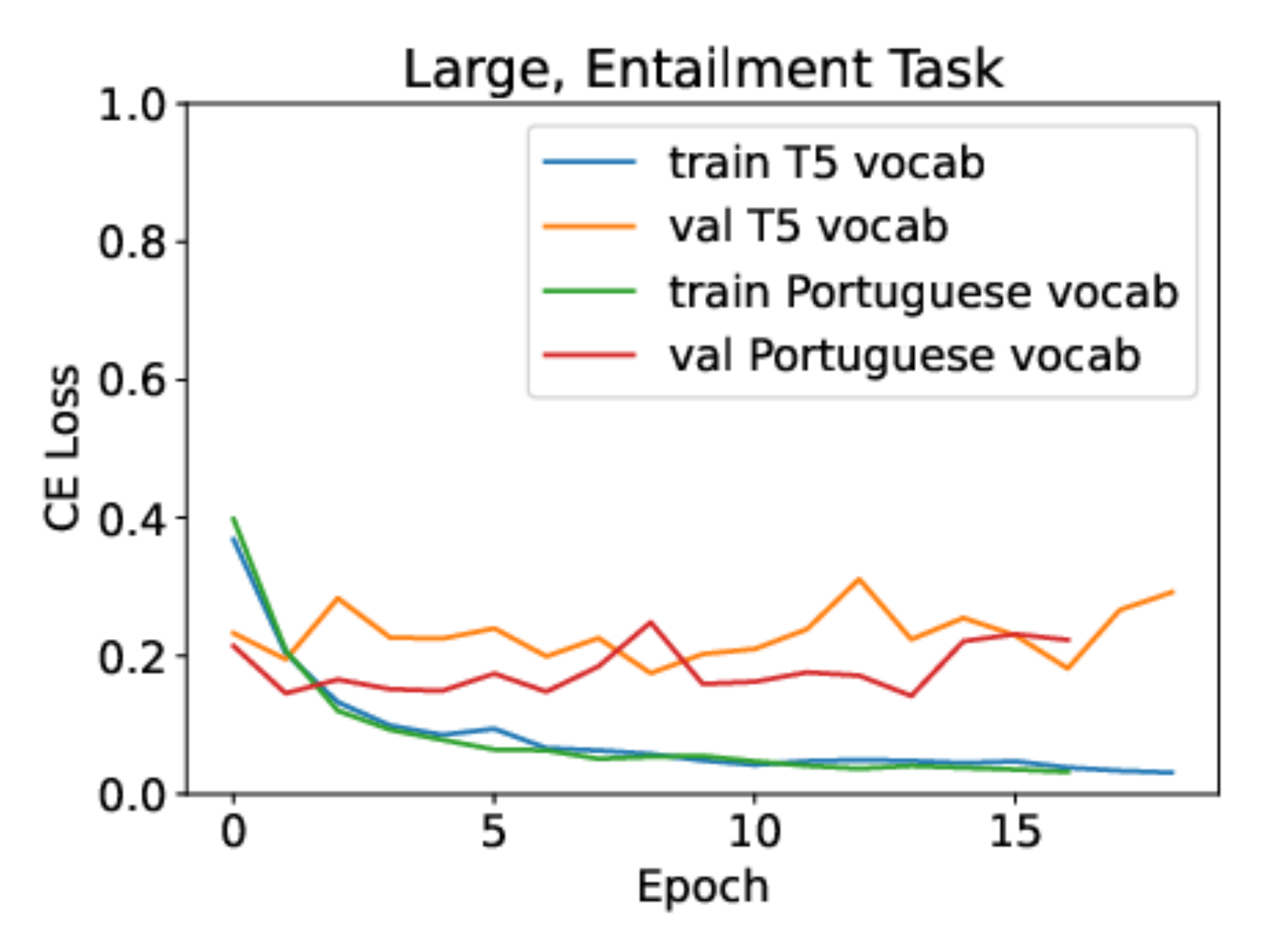}
  \caption{}
  \label{fig:large_vocab_vs_entail}
\end{subfigure}
\caption{Training and validation loss curves on ASSIN 2 tasks, starting from the pretrained PTT5 weights.}
\label{fig:vocab_vs}
\end{figure}

Figure~\ref{fig:vocab_vs} shows validation losses when varying the size of the initial PTT5 weights, with and without our custom Portuguese vocabulary on the semantic similarity task (Figures \ref{fig:small_vocab}, \ref{fig:base_vocab} and \ref{fig:large_vocab}) and entailment task (Figures~\ref{fig:vocab_vs_entail}, \ref{fig:base_vocab_vs_entail}, and \ref{fig:large_vocab_vs_entail}). We notice that the Portuguese vocabulary helps PTT5 achieving better results and convergence on ASSIN 2. Additionally, we notice that larger models converge faster in the fine-tuning step.

\subsubsection{Ablation: Additional Hyperparameter Tuning}

Table~\ref{tab:val} shows the best validation MSE loss for the similarity task and best validation cross-entropy loss for the entailment task, for each model. All PTT5 models were pretrained for 4 epochs. PTT5 base with the Portuguese vocabulary achieved the lowest MSE (0.0387). PTT5 Large with the Portuguese vocabulary achieved the lowest cross-entropy (0.1420). However, as shown in Table~\ref{tab:test}, the large model achieved a lower Person correlation than the base model.

\begin{table}[ht]
\centering
\resizebox{\columnwidth}{!}{%
\begin{tabular}{ccccccc}
\toprule
\textbf{Initial Weights} & \textbf{Size} & \textbf{Batch Size} & \textbf{LR} & \textbf{Vocab}  & \textbf{Val. MSE loss}  & \textbf{Val. CE loss} \\\midrule
T5  & Small & 32 & 1e-4 & T5                   & 0.1042     &  0.2697  \\
T5  & Base & 2 & 1e-4 & T5                    & 0.0626     &  0.1865  \\
T5  & Large & 1 & 1e-4 & T5                   & 0.0617     &  0.1732  \\
PTT5 & Small & 32 & 1e-4 & T5         & 0.0976     &  0.2135  \\
PTT5 & Base & 2 & 1e-4 & T5          & 0.0407     &  0.1700  \\
PTT5 & Large & 1 & 1e-4 & T5         & 0.0413     &  0.1750  \\
PTT5 & Small  & 32 & 1e-4 & PT         & 0.0551     &  0.1727      \\
PTT5 & Base & 2 & 1e-4 & PT          & \textbf{0.0387} & 0.1439 \\
PTT5 & Large & 1 & 1e-4 & PT         & 0.0460 &  \textbf{0.1420}
\\\bottomrule
\end{tabular}
}
\caption{\label{tab:val} Validation loss for each model configuration after fine-tuning on the ASSIN 2 tasks.  MSE loss is for the similarity task and cross-entropy loss (CE loss) is for the entailment task. Lower is better.}
\end{table}

In Table~\ref{tab:additional}, we show some additional experiments to explore variations in the batch size, learning rate, and the number of pretraining epochs. We also evaluated the performance of the PTT5 pretrained with vocabulary embeddings only.

\begin{table}[ht]
\centering
\resizebox{\columnwidth}{!}{%
\begin{tabular}{ccccccc}
\toprule
\textbf{Pretraining Epochs} & \textbf{Emb. Only?} & \textbf{Size} & \textbf{Batch Size} & \textbf{LR}  & \textbf{Val. MSE Loss} & \textbf{Val. Pearson}\\\midrule
4 & No & Base & 2 & 1e-4 & \textbf{0.0387} & \textbf{0.981} \\
4 & No & Base & 32 & 1e-4 & 0.0514 & 0.976\\
4 & No & Base & 32 & 1e-3 & 0.0483 & 0.975\\
4 & Yes & Base  & 2 & 1e-4 &  \textbf{0.0353} & \textbf{0.983}\\
4 & Yes & Base  & 32 & 1e-3 & 0.0447 & 0.978 \\
4 & No & Large & 1   & 1e-4 & 0.0378 & 0.981 \\
4 & No & Large & 2 & 1e-4 & 0.0393 & 0.980 \\
1 & No & Base   & 2 & 1e-4 & 0.0389 & 0.982 \\
0.1 & No & Base   & 2 & 1e-4 & 0.0367 & 0.982 \\
1 & Yes & Base   & 2 & 1e-4 & 0.04525 & 0.978 \\
0.1 & Yes & Base & 2 & 1e-4 & 0.0419 & 0.979 \\\bottomrule
\end{tabular}
}
\caption{Additional fine-tuning experiments on ASSIN 2's similarity task to explore learning rate, batch accumulation, and pretraining vocabulary embedding only. From PTT5 with Portuguese vocabulary checkpoint. PTT5 Large was pretrained with a batch size of 128.}
\label{tab:additional}
\end{table}

Different pretraining methods showed similar performance on the validation set of ASSIN 2. However, on the test set, pretraining all weights achieved far better performance than pretraining the vocabulary embeddings only. These results suggest that the validation set of ASSIN 2 was not the best choice for model selection, probably due to its small size (500 examples) and because metrics were already too high for all models (e.g., Pearson correlations were close to 1). Hence, experiments on other tasks are needed to confirm whether pretraining vocabulary embeddings only is a viable option. 

\subsection{HAREM Experiments}

Here we describe the experiments on the HAREM dataset for Named Entity Recognition. We compare three versions of the T5 models: the original T5 model pretrained on English texts, PTT5 with the original English vocabulary, and PTT5 with the Portuguese vocabulary. We use base models due to the computational costs of larger models. This constraint also permitted only a small batch size of 2, however we accumulate the the gradients over 4 steps, given a total batch size of 8. We utilize AdamW with a learning rate of $0.0002$ without any \textit{warmup} or \textit{scheduling} techniques. 

We preprocess First HAREM and MiniHAREM datasets using the code made available by Souza et al.~\cite{souza2019Portuguese}. For the models using the English vocabulary, we replace accented characters (e.g., ã, ó) with its closest non-accented representation (e.g., a, o). The entity tags are exactly their natural language labels. For instance, ``\textit{Organização}" is used to identify organizations. Due to the limitation of 512 tokens of the T5 model, all examples that are longer than 512 are divided into smaller segments using a sliding window technique and a stride of 256 tokens. 

During the validation phase, a \textit{beam search} decoding method of width 5 is used to generate the output sequence, followed by a token labeling postprocessing that uses the BIO format. For example, if the model generated ``John [Person] lives in [Other] New York [Local]'', we would mark the words in the sequence as B-PER, O, O, B-LOC, I-LOC. This format stands for Begin, Intermediate and Out of context tokens and allows us to easily compare with other works. However, as a drawback, any non-entity token inserted or forgotten on the output that does not exist on the input sentence can lead to misalignment of the predicted and true labels sequences, thus impacting performance negatively.

The results are shown in Table~\ref{tab:harem}. Our language-specific pre-training (PTT5) helps improving upon the original T5 pretraining, with a slight gain when a Portuguese vocabulary is used. Nevertheless, Portuguese BERT still performs better than PTT5 by a small amount.

Table~\ref{tab:harem_ents} shows that most degrading point of the performance comes from detecting the Value entity.

\begin{table}[ht]
\centering
\begin{tabular}{ccccc}
\toprule
\textbf{Architecture} & \textbf{Vocab}  & \textbf{Precision}  & \textbf{Recall} & \textbf{F1} \\\midrule
T5 Base  & T5 & 71.7 & 71.2 & 71.5 \\
PTT5 Base  & T5 & 78.1 & 79.0 & 78.5  \\
PTT5 Base  & PT & \textbf{85.5} & 78.8 & 82.0 \\
BERTimbau Base~\cite{souza2019Portuguese} & PT & 81.9 & \textbf{82.7} & \textbf{82.2}\\
\bottomrule
\end{tabular}
\caption{\label{tab:harem} Results on the MiniHAREM dataset on the \textit{selective} scenario.}
\end{table}

\begin{table}[ht]
\centering
\begin{tabular}{cccc}
\toprule
\textbf{Entity} & \textbf{Precision}  & \textbf{Recall}  & \textbf{F1} \\\midrule
Time  & 87.2 & 84.1 & 85.6 \\
Local & 87.6 & 82.3 & 84.9  \\
Person  & 86.8 & 79.2 & 82.8 \\
Organization & 80.2 & 78.4 & 79.3 \\
Value & 84.4 & 63.2 & 72.3 \\
\bottomrule
\end{tabular}
\caption{\label{tab:harem_ents} Results on the MiniHAREM entities of the \textit{selective} scenario by the PTT5 with Portuguese vocab}
\end{table}

\section{Conclusion}
We pretrained T5 models on a large Brazilian Portuguese corpus. The resulting models achieved better performance than the original T5 models on the Portuguese sentence entailment and NER tasks. Moreover, using a Portuguese vocabulary proved to be better than using the original T5 vocabulary. Finally, for ASSIN 2 tasks, we found that pretraining all weights leads to better performance than pretraining the vocabulary embeddings only. Despite having achieved better results than the top-submissions to the ASSIN 2 leaderboard, our Portuguese T5 model is still a few points below to the state-of-the-art model, a Portuguese BERT Large model (BERTimbau Large).

\section*{Acknowledgements}
We thank Google for the free TPUs and Google Cloud credits. This work was initially developed as the final project for the IA376E course taught by professors Rodrigo Nogueira and Roberto Lotufo at the University of Campinas (UNICAMP).

\bibliographystyle{plain}
\bibliography{main.bib}

\begin{thebibliography}{10}

\bibitem{baly2020arabert}
Fady Baly, Hazem Hajj, et~al.
\newblock Arabert: Transformer-based model for arabic language understanding.
\newblock In {\em Proceedings of the 4th Workshop on Open-Source Arabic Corpora
  and Processing Tools, with a Shared Task on Offensive Language Detection},
  pages 9--15, 2020.

\bibitem{Canete2020beto}
José Cañete, Gabriel Chaperon, Rodrigo Fuentes, and Jorge Pérez.
\newblock Spanish pre-trained bert model and evaluation data.
\newblock In {\em to appear in PML4DC at ICLR 2020}, 2020.

\bibitem{delobelle2020robbert}
Pieter Delobelle, Thomas Winters, and Bettina Berendt.
\newblock {RobBERT: a Dutch RoBERTa-based Language Model}.
\newblock {\em arXiv preprint arXiv:2001.06286}, 2020.

\bibitem{devlin2018bert}
Jacob Devlin, Ming-Wei Chang, Kenton Lee, and Kristina Toutanova.
\newblock Bert: Pre-training of deep bidirectional transformers for language
  understanding.
\newblock {\em arXiv preprint arXiv:1810.04805}, 2018.

\bibitem{fonseca2020assin2-stilingue}
Evandro Fonseca and Jo{\~a}o Paulo~Reis Alvarenga.
\newblock Wide and deep transformers applied to semantic relatedness and
  textual entailment.
\newblock In Oliveira et~al. \cite{ASSIN2}, pages 68--76.

\bibitem{kudo2018subword}
Taku Kudo.
\newblock Subword regularization: Improving neural network translation models
  with multiple subword candidates.
\newblock {\em arXiv preprint arXiv:1804.10959}, 2018.

\bibitem{kudo2018sentencepiece}
Taku Kudo and John Richardson.
\newblock Sentencepiece: A simple and language independent subword tokenizer
  and detokenizer for neural text processing.
\newblock {\em arXiv preprint arXiv:1808.06226}, 2018.

\bibitem{kuratov2019rubert}
Yuri Kuratov and Mikhail Arkhipov.
\newblock Adaptation of deep bidirectional multilingual transformers for
  russian language.
\newblock {\em arXiv preprint arXiv:1905.07213}, 2019.

\bibitem{liu2019variance}
Liyuan Liu, Haoming Jiang, Pengcheng He, Weizhu Chen, Xiaodong Liu, Jianfeng
  Gao, and Jiawei Han.
\newblock On the variance of the adaptive learning rate and beyond.
\newblock {\em arXiv preprint arXiv:1908.03265}, 2019.

\bibitem{liu2019roberta}
Yinhan Liu, Myle Ott, Naman Goyal, Jingfei Du, Mandar Joshi, Danqi Chen, Omer
  Levy, Mike Lewis, Luke Zettlemoyer, and Veselin Stoyanov.
\newblock Roberta: A robustly optimized bert pretraining approach.
\newblock {\em arXiv preprint arXiv:1907.11692}, 2019.

\bibitem{loshchilov2017decoupled}
Ilya Loshchilov and Frank Hutter.
\newblock Decoupled weight decay regularization, 2017.

\bibitem{martin2019camembert}
Louis Martin, Benjamin Muller, Pedro Javier~Ortiz Su{\'a}rez, Yoann Dupont,
  Laurent Romary, {\'E}ric~Villemonte de~la Clergerie, Djam{\'e} Seddah, and
  Beno{\^\i}t Sagot.
\newblock Camembert: a tasty french language model.
\newblock {\em arXiv preprint arXiv:1911.03894}, 2019.

\bibitem{nguyen2020phobert}
Dat~Quoc Nguyen and Anh~Tuan Nguyen.
\newblock Phobert: Pre-trained language models for vietnamese.
\newblock {\em arXiv preprint arXiv:2003.00744}, 2020.

\bibitem{ASSIN2}
Hugo~Gonçalo Oliveira, Livy Real, and Erick Fonseca, editors.
\newblock {\em Proceedings of the ASSIN 2 Shared Task: Evaluating Semantic
  Textual Similarity and Textual Entailment in Portuguese, {Extended}
  {Semantic} {Web} {Conference}}, number 2583 in CEUR Workshop Proceedings,
  2020.

\bibitem{polignano2019alberto}
Marco Polignano, Pierpaolo Basile, Marco de~Gemmis, Giovanni Semeraro, and
  Valerio Basile.
\newblock {AlBERTo: Italian BERT Language Understanding Model for NLP
  Challenging Tasks Based on Tweets}.
\newblock In {\em Proceedings of the Sixth Italian Conference on Computational
  Linguistics (CLiC-it 2019)}, volume 2481. CEUR, 2019.

\bibitem{raffel2019exploring}
Colin Raffel, Noam Shazeer, Adam Roberts, Katherine Lee, Sharan Narang, Michael
  Matena, Yanqi Zhou, Wei Li, and Peter~J Liu.
\newblock Exploring the limits of transfer learning with a unified text-to-text
  transformer.
\newblock {\em arXiv preprint arXiv:1910.10683}, 2019.

\bibitem{real2020assin}
Livy Real, Erick Fonseca, and Hugo~Gon{\c{c}}alo Oliveira.
\newblock The assin 2 shared task: a quick overview.
\newblock In {\em International Conference on Computational Processing of the
  Portuguese Language}, pages 406--412. Springer, 2020.

\bibitem{rodrigues2020assin2-dlb}
Ruan Rodrigues, J\'{e}ssica da~Silva, Pedro Castro, Nadia Felix, and Anderson
  Soares.
\newblock Multilingual transformer ensembles for portuguese natural language
  tasks.
\newblock In Oliveira et~al. \cite{ASSIN2}, pages 27--38.

\bibitem{rodrigues2020assin2-ipr}
Rui Rodrigues, Paula Couto, and Irene Rodrigues.
\newblock Ipr: The semantic textual similarity and recognizing textual
  entailment systems.
\newblock In Oliveira et~al. \cite{ASSIN2}, pages 39--47.

\bibitem{santos-etal-2006-harem}
Diana Santos, Nuno Seco, Nuno Cardoso, and Rui Vilela.
\newblock {HAREM}: An advanced {NER} evaluation contest for {P}ortuguese.
\newblock In {\em Proceedings of the Fifth International Conference on Language
  Resources and Evaluation ({LREC}{'}06)}, Genoa, Italy, May 2006. European
  Language Resources Association (ELRA).

\bibitem{shazeer2018adafactor}
Noam Shazeer and Mitchell Stern.
\newblock Adafactor: Adaptive learning rates with sublinear memory cost.
\newblock {\em arXiv preprint arXiv:1804.04235}, 2018.

\bibitem{souza2019Portuguese}
F{\'a}bio Souza, Rodrigo Nogueira, and Roberto Lotufo.
\newblock Portuguese named entity recognition using bert-crf.
\newblock {\em arXiv preprint arXiv:1909.10649}, 2019.

\bibitem{speer-2019-ftfy}
Robyn Speer.
\newblock ftfy.
\newblock Zenodo, 2019.
\newblock Version 5.5.

\bibitem{virtanen2019multilingual}
Antti Virtanen, Jenna Kanerva, Rami Ilo, Jouni Luoma, Juhani Luotolahti, Tapio
  Salakoski, Filip Ginter, and Sampo Pyysalo.
\newblock Multilingual is not enough: Bert for finnish.
\newblock {\em arXiv preprint arXiv:1912.07076}, 2019.

\bibitem{vries2019bertje}
Wietse~de Vries, Andreas~van Cranenburgh, Arianna Bisazza, Tommaso Caselli,
  Gertjan~van Noord, and Malvina Nissim.
\newblock {BERTje}: {A} {Dutch} {BERT} {Model}.
\newblock {\em arXiv preprint arXiv:1912.09582}, December 2019.

\bibitem{wagner2018brwac}
Jorge~A Wagner~Filho, Rodrigo Wilkens, Marco Idiart, and Aline Villavicencio.
\newblock The brwac corpus: A new open resource for brazilian portuguese.
\newblock In {\em Proceedings of the Eleventh International Conference on
  Language Resources and Evaluation (LREC 2018)}, 2018.

\end{thebibliography}

\end{document}